 \documentclass[pmlr,twocolumn,10pt]{jmlr} 

 \usepackage{rotating}
 \usepackage{longtable}

\usepackage{booktabs}
\usepackage{siunitx}
\usepackage{algorithm}
\usepackage{algpseudocode}
\usepackage{multirow}
\usepackage{adjustbox}
\usepackage{pifont}

\theorembodyfont{\upshape}
\theoremheaderfont{\scshape}
\theorempostheader{:}
\theoremsep{\newline}

\jmlrvolume{225}
\jmlryear{2023}
\jmlrsubmitted{LEAVE UNSET}
\jmlrpublished{LEAVE UNSET}
\jmlrworkshop{Machine Learning for Health (ML4H) 2023} 

 \title[Gradient-Map-Guided Adaptive Domain Generalization for Cross Modality MRI Segmentation]{Gradient-Map-Guided Adaptive Domain Generalization for Cross Modality MRI Segmentation}

\author{%
\Name{Bingnan Li} \Email{libn@shanghaitech.edu.cn}\\
\addr School of Information Science and Technology, ShanghaiTech University
\AND
\Name{Zhitong Gao}
\Email{gaozht@shanghaitech.edu.cn}\\
\addr School of Information Science and Technology, ShanghaiTech University
\AND
\Name{Xuming He} \Email{hexm@shanghaitech.edu.cn}\\
\addr School of Information Science and Technology, ShanghaiTech University\\
\addr Shanghai Engineering Research Center of Intelligent Vision and Imaging
}

\begin{document}

\maketitle

\begin{abstract}
Cross-modal MRI segmentation is of great value for computer-aided medical diagnosis, enabling flexible data acquisition and model generalization. However, most existing methods have difficulty in handling local variations in domain shift and typically require a significant amount of data for training, which hinders their usage in practice. To address these problems, we propose a novel adaptive domain generalization framework, which integrates a learning-free cross-domain representation based on image gradient maps and a class prior-informed test-time adaptation strategy for mitigating local domain shift. We validate our approach on two multi-modal MRI datasets with six cross-modal segmentation tasks. Across all the task settings, our method consistently outperforms competing approaches and shows a stable performance even with limited training data. 
Our Codes are available now at \url{https://github.com/cuttle-fish-my/GM-Guided-DG}.
\end{abstract}

\section{Introduction}
\label{sec:Introduction}
Semantic segmentation of Magnetic Resonance Imaging (MRI) sequences, a core task in computer-aided medical diagnoses, has achieved remarkable progress due to the powerful representation learning based on deep neural networks~\citep{ronneberger2015u, isensee2021nnu}. Conventional approaches typically assume the same data distribution and share the same modality setting in both training and test stages~\citep{zhou2019review}. Despite their promising results, these methods either employ multi-modal inputs or learn a separate model for each modality, which induces high time or financial costs. To alleviate this, one promising strategy involves training a model on one specific modality and enabling it to adapt to other modalities during the test phase, which is referred to as Single Domain Generalization (SDG)~\citep{qiao2020learning}. However, it is challenging to apply the common SDG strategy to MRI segmentation due to the large domain gaps between different modalities and often limited training data in medical applications.
\begin{figure}[tbp]
  \label{fig:AD_img}
  {%
    \includegraphics[width=\linewidth]{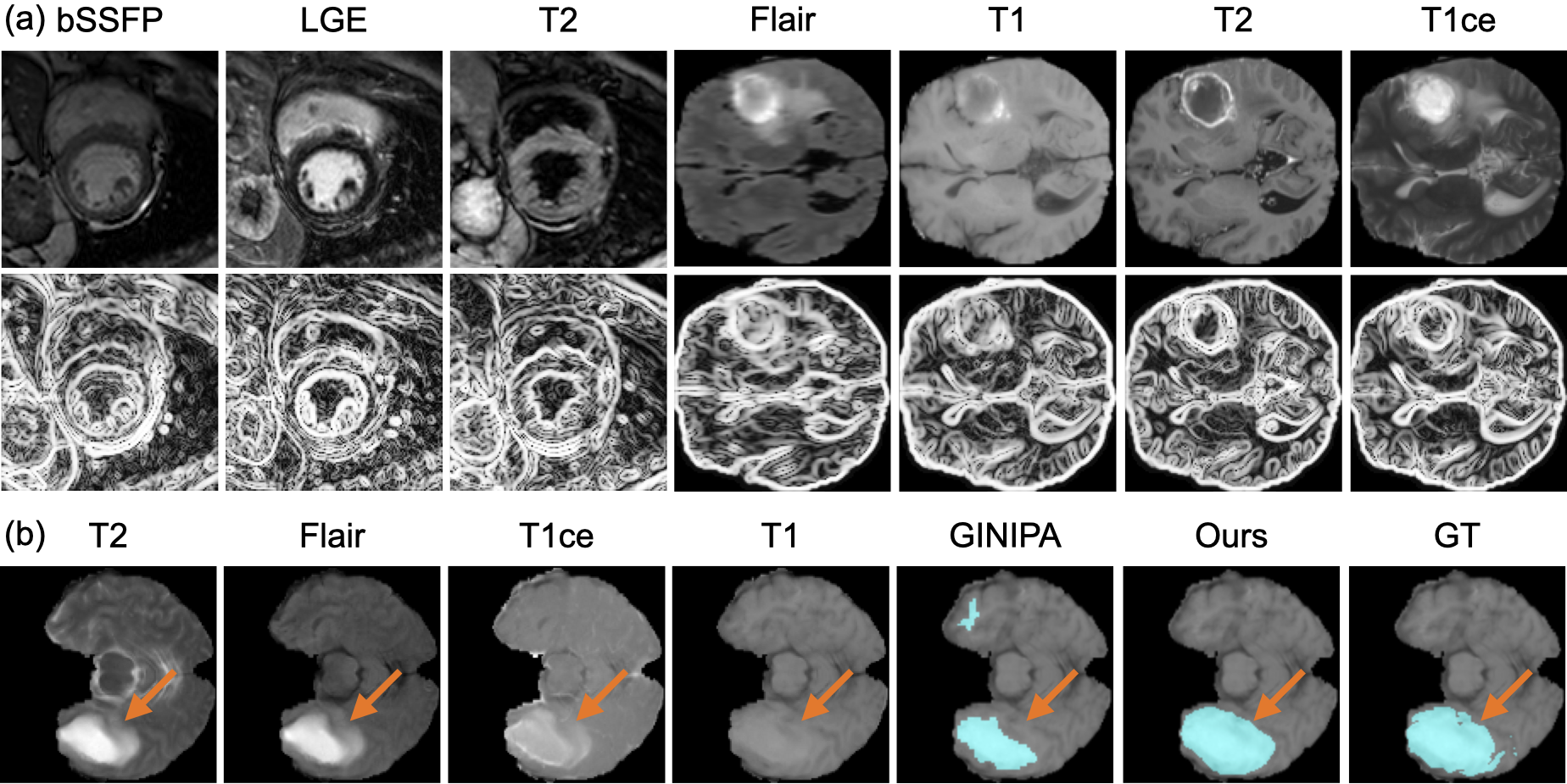}
  }
  \caption{(a) An illustration of original images and Gradient-Map Representation (GMR) which can effectively mitigate the domain gap; 
  (b) Comparison of segmentation outputs (GINIPA vs. our method) when local variations exist for different modalities.
}
\end{figure}
There have been a few attempts to tackle the problem of SDG for MRI image segmentation, which mostly focus on learning a domain-invariant feature space by designing specific learning strategies~\citep{liu2020shape, xu2022adversarial} or domain randomization techniques~\citep{zhang2020generalizing, ouyang2022causality, liu2021disentangled}.

Nonetheless, the efficacy of these learned feature spaces is highly contingent on the amount of training data and their performance tends to degrade rapidly with fewer data (see \figureref{fig:efficiency} for details). Moreover, most of these methods focus on generalization with image-level domain variation and may struggle with local variations in different domains, such as the different responses of lesion tissues across modalities. More recent work~\citep{liu2022single} introduces Test-Time Adaptation (TTA) techniques coupled with a shape dictionary representation. However, such a strategy requires spatial alignment of target regions, and hence is limited in handling objects with diverse shapes and locations (e.g., brain tumors~\citep{menze2014multimodal}).

In this work, we aim to address the aforementioned limitations by introducing a novel gradient-map-guided adaptive domain generalization framework for MRI segmentation. In particular, we adopt a learning-free domain-invariant input representation based on the image gradient map, which is capable of mitigating the global image style shift across different modalities. Based on that, we then develop a new test-time adaption strategy that integrates semantic class prior with pseudo-label-based self-training. This allows us to cope with local appearance variations and adapt to image-specific pixel-wise class distributions.   

Specifically, we adopt a two-phase domain generalization procedure for cross-modal MRI segmentation. During the training phase, our method encodes input images into their gradient maps (see \figureref{fig:AD_img} (a) for an illustration), and trains a semantic segmentation network based on the training data of the source modality. During the test phase, we further fine-tune the segmentation model on each input image from the target modality in an iterative manner using the pseudo-labels of pixels generated with a dynamically-estimated pixel-wise class prior. 

We evaluate our method on two multi-modality MRI datasets, including MS-CMRSeg2019~\citep{zhuang2018multivariate} and BraTS2018~\citep{menze2014multimodal}, under the standard and limited training data settings. The results show the superiority of our method in comparison with recent domain generalization and TTA methods. We summarize our contributions as follows:
\begin{itemize}
    \item We propose an adaptive single domain generalization method for cross-modal semantic segmentation of MRI images to better cope with both global and instance-specific domain gaps.
    \item We introduce an efficient gradient-based domain-invariant representation in MRI image segmentation and develop an effective TTA strategy based on class prior-aware self-training.
    \item Our method achieves state-of-the-art performance on cross-modal cardiac segmentation and brain tumor segmentation tasks.
\end{itemize}

\section{Related Work}
\label{sec:RelatedWork}
The cross-modality MRI segmentation problem has received significant attention in the literature, and a variety of approaches have been developed under different problem settings. A straightforward solution is to adapt a source-domain-trained model to the target domain during training, which is referred to as Unsupervised Domain Adaptation (UDA)~\citep{chen2020unsupervised, vesal2021adapt, yang2019unsupervised}. However, the requirement for target data during training poses a practical challenge, as acquiring target data can be arduous or infeasible in many real-world scenarios. This limitation has led to the widespread use and growing importance of Domain Generalization (DG) and Test-Time Adaptation (TTA) methods, which only require access to source domain data during the training stage. In the subsequent sections, we offer an overview of the relevant work within the domains of DG and TTA.



\paragraph{Domain Generalization}
The goal of Domain Generalization is to learn a model capable of generalizing to any other domain during the test. While some techniques rely on training data from multiple domains~\citep{pereira2016brain, dolz2018hyperdense}, we focus on a more challenging setting where only one domain is available during training. This is also called Single Domain Generalization (SDG). 

Existing methods address the problem by encouraging models to learn a domain-invariant feature space implicitly. For instance, domain randomization methods like random weighted network~\citep{ouyang2022causality} and data augmentations~\citep{zhang2020generalizing, otalora2019staining,chen2020improving} simulate potential target domain information, promoting the development of domain-robust models. In addition to these techniques, special learning strategies like meta-learning~\citep{oliveira2022domain, liu2020shape}, disentangle learning~\citep{liu2021disentangled} and adversarial learning~\citep{xu2022adversarial} are used to regularize the model for domain invariance. 
However, most of these methods focus on generalization on global-level domain shifts, such as different styles, and struggle with local appearance variations across different domains. Besides, the efficacy of these learned feature spaces could also be highly contingent on the size of the training dataset. 
By contrast, we utilize a learning-free gradient-map-based domain-invariant representation to handle the global style variation.
During test time, we introduce a pseudo-label-based self-training strategy to mitigate the local domain shifts that can vary inconsistently across different modalities.

\paragraph{Test Time Adaptation}
The objective of Test-Time Adaptation (TTA) is to harness online test data to adapt the model during the testing phase. Several techniques have been explored in this field.~\citet{wang2020tent} introduce a novel method that employs entropy minimization exclusively on BathNorm (BN) layers to force a confident prediction.~\citet{he2021autoencoder} utilize Auto Encoder (AE) as a domain distance indicator and update the model via the reconstruction loss.~\citet{wang2023dynamically} introduces prototype-based methods into the realm of TTA and harness the feature distance to different prototype as a measure of classification probability.~\citet{tang2023neuro} proposed a BP-free TTA method using Hebbian layers that inspired from Hebbian learning~\citep{hebb2005organization}. Nevertheless, TTA methods often rely heavily on the base model performance, and some require special network designs such as AE and Hebbian layers. Recently,~\citet{liu2022single} introduced a novel approach that combines DG and TTA by incorporating dictionary learning during training and constraining the consistency of the dictionary coefficient between two different noise-disturbed images in test time. However, the strong assumption about shape-invariant and location-invariant segmentation targets across source and target domains limits the method when handling objects with diverse shapes and locations.

In contrast, our proposed pseudo-label-based TTA strategy utilizes class-prior to prioritize pseudo-label generation of the foreground classes. This approach empowers our models to produce high-quality segmentations, even in cases where the initial predictions are far from ideal. Besides, our method avoids assumptions about the shape and location of the segmentation target, which demonstrates the flexibility and applicability of our proposed method.

\section{Methodology}
\label{sec:Methodology}
In this section, we introduce our adaptive domain generalization framework for MRI segmentation, which proposes two novel components to cross-modality segmentation network learning: 1) a modality-robust input representation based on gradient maps and 2) a class prior-informed test-time adaptation strategy. An overview of our framework is shown in \figureref{fig:pipeline}.  
Below, we first introduce the problem setting in \sectionref{sec:ProblemFormulation}, then describe the design of our gradient-map-based semantic segmentation network and its training on the source domain in \sectionref{sec:GradientMapRepresentation}. This is followed by our prior-informed test time adaptation in \sectionref{sec:PriorInformedAdapatation}.

\begin{figure*}[htbp]
    \floatconts
    {fig:pipeline}
    {\caption{Overview of our method. Subgraph (a) and (b) refer to the training and test stage pipeline respectively. For the training state, we employ the gradient map $G(\mathbf{x})$ as a domain-invariant representation to mitigate image-level domain shift. For test stage, we utilize class prior $\mathbf{p}$ to adapt posterior $P(\mathbf{y}|\mathbf{x})$ for N times, alleviating local discrepancy.}}
    {\includegraphics[width=0.9\linewidth]{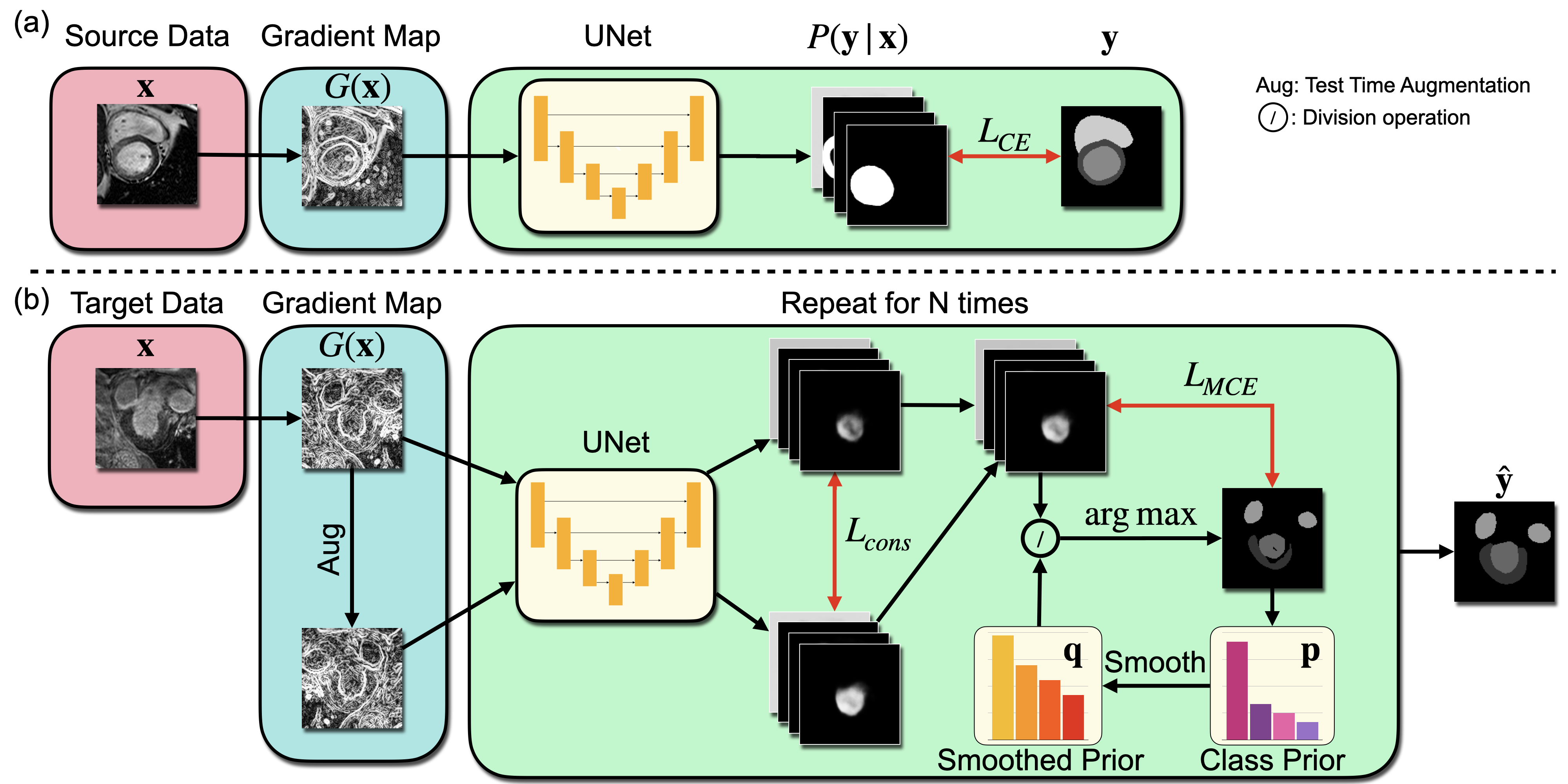}}
\end{figure*}

\subsection{Problem Setting}
\label{sec:ProblemFormulation}
Consider a training dataset $D = \{\mathbf{x}_n,\mathbf{y}_n\}_{n=1}^N$ with $\{\mathbf{x}_n,\mathbf{y}_n\}\sim P^s_{XY}$, where $\mathbf{x}_n \in R^{H\times W}$ is a 2D MRI image from a source modality and $\mathbf{y}_n \in [1,C]^{H\times W}$ is the corresponding segmentation label. Here $H,W$ are the spatial dimensions and $C$ is the number of semantic classes. The goal of cross-modality semantic segmentation is to learn a segmentation model $\mathcal{M}_\theta$ from $D$ that can be generalized to images from any other modalities during the test. 
In this paper, we aim to design an adaptive domain generalization framework, which utilizes both the source-modality data and the online unlabeled test data to adapt the network to fit the target modality, where the test image $\mathbf{x} \sim P^t_{XY}$ with $P^t_{X}\neq P^s_{X}$. 

\subsection{Modality-robust Segmentation Network}
\label{sec:GradientMapRepresentation}
We adopt a U-Net~\citep{ronneberger2015u} as our segmentation backbone, and first build a modality-robust model using the source-modality dataset $D$. In particular, as detailed below, we introduce an image gradient map-based representation as the network input and a heavy data augmentation strategy in model training, both of which mitigate the modality bias in the learned segmentation network.    

\paragraph{Gradient-Map Representation}
We first introduce our \textit{Gradient-Map Representation (GMR)} as a modality-invariant input representation, which leverages the image prior to achieve modality generalization with high data efficiency.  
Our design is inspired by a key observation that while different modalities may have unique global styles, the most pronounced variations in intensity occur at relatively consistent locations, typically at the boundaries between adjacent tissues. This observation motivates us to use the gradient map, which captures these relative intensity changes (structural information), as a modality-robust representation.

Formally, the gradient map of an input image $\mathbf{x}$, denoted as $G(\mathbf{x})$, is computed as follows:
\begin{equation}
    \label{eq:GMR}
    G(\mathbf{x})=HE\left(\sqrt{corr\left(\mathbf{x}, \mathbf{K}\right)^2 + corr\left(\mathbf{x},\mathbf{K}^T\right)^2}\right),
\end{equation}
where $HE(\cdot)$ refers to the histogram equalization, $corr(\cdot, \cdot)$ represents the correlation operation and $\mathbf{K}=[-1\; 0\; 1]$.
We visualize GMR examples in \figureref{fig:AD_img} (a), which shows that the discrepancy between different modalities has been effectively reduced and the structural information is well preserved. 

\paragraph{Model Training}
We employ the Heavy Augmentation~\citep{vesal2021adapt} technique in the training process. This method involves augmenting both the source image $\mathbf{x}$ and its corresponding annotation $\mathbf{y}$ using a transformation $T\in \mathcal{T}$ sampled from a distribution $P(T)$. 
The optimization problem of the training process can be described as follows:
\begin{align}
    \mathop{min}\limits_{\theta} \mathbb{E}_{T}\left[L_{ce}\left(\mathcal{M}_{\theta}\left(T^{(X)}\left(G\left(\mathbf{x}\right)\right)\right), T^{(Y)}(\mathbf{y})\right)\right]
\end{align}
where $L_{ce}(\cdot, \cdot)$ is the Cross-Entropy (CE) loss and $T^{(X)/(Y)}(\cdot)$ refer to the data augmentation function applied to input data $\mathbf{x}$ and annotation $\mathbf{y}$.

\subsection{Prior-Informed Test-Time Adaptation}
\label{sec:PriorInformedAdapatation}

While the GRM representation reduces the modality gap in terms of global intensity style shift, it is unable to eliminate modality-specific local appearance variations. For instance, \figureref{fig:AD_img} (b) displays a brain MRI with arrows indicating the local intensity variation that is inconsistent across four modalities. Such modality-specific variation hinders the model from generalizing across different modalities. 
 
To tackle this, we propose a test-time adaptation strategy to finetune the segmentation network with unlabeled test data in an online fashion. In particular, we develop a self-training method that gradually expands the regions of certain classes, which typically include the regions with modality-specific variation (e.g., lesion areas), via an iterative self-labeling scheme. We refer to this process as \textit{Prior-Informed Test Time Adaptation (PITTA)}. 

Specifically, given a test image $\mathbf{x}$, we start from the model trained on the source modality as in ~\sectionref{sec:GradientMapRepresentation} and perform $N$ adaptation iterations. For each iteration, we first merge the BatchNorm (BN) statistics from the source and target domain to stabilize prediction~\citep{Wang_2023_CVPR}, and then employ consistency loss and Masked Cross Entropy (MCE) loss to dynamically adjust the model. Below we will introduce the details of each step in PITTA.



\paragraph{Test Time Augmentation}
Instead of direct prediction, we leverage Test Time Augmentation~\citep{zeng2017deepem3d,ronneberger2015u,beier2017multicut} to improve the robustness of model prediction. 
In detail, for the $n$-th iteration, we utilize a test-time augmentation function $A_T(\cdot)$\footnote{Note that our test-time augmentation $A_T(\cdot)$ differs from the Heavy Augmentation in model training (\sectionref{sec:GradientMapRepresentation}). to augment the input MRI, and} generate two logit predictions from the original input and its augmented version, respectively. We denote those two outputs as $\mathbf{S}^{(n)}$ and $\mathbf{S}'^{(n)}$, which are computed as follows:
\begin{align}
    \mathbf{S}^{(n)} &= \mathcal{M}_{\theta}(G(\mathbf{x}))\\
    \mathbf{S}'^{(n)} &= A_T^{-1}(\mathcal{M}_{\theta}(A_T(G(\mathbf{x}))))
\end{align}
where 
$A_T^{-1}(\cdot)$ refers to the inverse augmentation function, which serves to counteract the initial spatial augmentations. We take the average of the two logit outputs to generate the pixel-wise class probabilities as follows,
\begin{align}
    \label{eq:posterior} {P^{(n)}(\mathbf{y}|\mathbf{x})} &= \mathop{softmax}\left(\left(\mathbf{S}^{(n)}+\mathbf{S}'^{(n)}\right)/2\right).
\end{align}
\paragraph{Prior-Informed Self-training}
We now introduce our self-training strategy to finetune the model on the test data. In each iteration, our training loss consists of two terms: one is a consistency loss $L_{cons}$ and the other is a pseudo-label-based MSE loss $L_{mce}$. We fine-tune the segmentation network with the following total loss:
\begin{equation}
    L_{total} = L_{mce} + \lambda L_{cons}
\end{equation}
where $\lambda$ is the weighting coefficient. Below we describe the details of two losses in the $n$-th iteration. 

\textit{1) Consistency Loss.} Given the two outputs $\mathbf{S}^{(n)}$,$\mathbf{S}'^{(n)}$, we define the consistency loss as below: 
\begin{align}
    L_{cons} &= |\mathbf{S}^{(n)}-\mathbf{S}'^{(n)}|.
\end{align}

\textit{2) Pseudo label-based Loss.} We design an adaptive pseudo-label generation scheme that utilizes the class prior information to better cope with the local variations across modalities and the distinctive class distribution of each input. Specifically, we first estimate a per-image class distribution $\mathbf{p}^{(n)}$ from the class distribution of the training set and the model prediction on the current test input. Based on this estimation, we then perform a re-weighting on prediction probability to increase the proportion of pseudo labels from smaller classes with less data. 

Such a scheme improves the model adaptation from two aspects: First, we observe that cross-modal local variations are typically associated with foreground classes with smaller regions. Our re-weighting enables the model fine-tuning to focus on those classes by generating more training pixels with pseudo labels from them. Second, by gradually shifting from the training class prior to (estimated) per-image class distribution, our model is able to adapt to each test image by properly re-balancing the probability scores of all the classes.   

Formally, we initialize $\mathbf{p}$ as the empirical class distribution estimated from the training set in the first iteration (i.e., $n=1$). 
Denote the number of classes as $C$, and at the $n$-th iteration, we compute a re-weighting probability vector $\mathbf{q}^{(n)}$ from the estimated per-image class distribution $\mathbf{p}^{(n)}$ as follows:
\begin{align}
    \label{eq:smoothpriror} \mathbf{q}^{(n)} = \alpha\frac{1}{C}\mathbf{1} + (1-\alpha) \mathbf{p}^{(n)}
\end{align}
where $\alpha$ is the coefficient for smoothing and $\mathbf{1}$ is a full-one vector of length $C$. We then generate our pseudo labels for the input $\mathbf{x}$ using a re-weighted probability output as below:
\begin{align}
    \hat{\mathbf{y}}_{ij}^{(n)} = \arg\max\left({P^{(n)}(\mathbf{y}_{ij}|\mathbf{x})}/{\mathbf{q}^{(n)}}\right)
\end{align}
where $ij$ is the pixel index and $/$ denotes element-wise division of two vectors.  

Given the generated pseudo labels $\mathbf{\hat{y}}^{(n)}$, we define our pseudo-label-based loss using a masked version of the Cross-Entropy (MCE). This choice is motivated by the fact that cross-modal local variations typically fall within a subset of classes, and the classes that are already well-segmented provide little loss feedback. Specifically, denote the subset as $\mathcal{C}_{lv}$ and the corresponding pixel mask as $\mathbf{M}$, we introduce our MCE loss term as follows,
\begin{align}
    L_{mce} &= L_{ce}\left(\mathbf{M} \odot P^{(n)}(\mathbf{y}|\mathbf{x}), \mathbf{\hat{y}}^{(n)}\right)
\end{align}
where $\mathbf{M}_{i,j} = \bigcup_{c\in \mathcal{C}_{lv}}\mathbf{1}\left(\hat{\mathbf{y}}^{(n)}_{i,j}=c\right)$, $\mathbf{1}(\cdot)$ is the indicator function and $\odot$ is the element-wise multiplication.

After the model update, we compute the class frequency from the model output $\mathbf{\hat{y}}^{(n)}$ on the test image and update the per-image class distribution $\mathbf{p}$ based on the Exponential Moving Average (EMA) as below:
\begin{align}
    \mathbf{p}^{(n+1)} = \alpha \mathbf{p}^{(n)} + (1-\alpha)\#\left\{\mathbf{\hat{y}}^{(n)}\right\}
\end{align}
where $\alpha$ is the weight coefficient and  $\#\{\cdot\}$ indicates the function computing the class frequency from the pseudo labels.  
After the $N$-th iteration, the model outputs the final prediction $\mathbf{\hat{y}}$ as follows:
\begin{align}
    \hat{\mathbf{y}}_{ij} = \arg\max\left({P^{(N)}(\mathbf{y}_{ij}|\mathbf{x})}/{\mathbf{q}^{(N)}}\right)
\end{align}
Note that we use re-weighted probability scores in order to take into account per-image class imbalance. The pseudo-code of our Prior-Informed Adaptation process is shown in~\appendixref{sec:pseudocode}~\algorithmref{alg:pseduocode}.

\begin{table*}[h]
    \floatconts
    {tab:comparison_MSCMRSeg2019}%
    {\caption{Quantitative comparisons in terms of Volumetric Dice scores on MS-CMRSeg2019 dataset.}}%
    {
    \begin{tabular}{l c c c c c}
            \toprule
            Setting & & \multicolumn{4}{c}{bSSFP$\to$T2} \\
            \midrule 
            Method & Category & Myo & LV & RV & Average\\
            \midrule
            
            SrcOnly & - & 0.0712{\scriptsize $\pm$ 0.0386} & 0.0687{\scriptsize $\pm$ 0.0490} & 0.0068{\scriptsize $\pm$ 0.0093} & 0.0489{\scriptsize $\pm$ 0.0302} \\
                        
            SAM++ & - & 0.8364{\scriptsize $\pm$ 0.0004} & 0.8370{\scriptsize $\pm$ 0.0014} & 0.7619{\scriptsize $\pm$ 0.0060} & 0.8118{\scriptsize $\pm$ 0.0017} \\
                        
            MinEnt & - & 0.0482{\scriptsize $\pm$ 0.0067} & 0.0513{\scriptsize $\pm$ 0.0051} & 0.0153{\scriptsize $\pm$ 0.0108} & 0.0382{\scriptsize $\pm$ 0.0067} \\
            
            \midrule
            
            HA & DG & 0.6600{\scriptsize $\pm$ 0.0064} & 0.7576{\scriptsize $\pm$ 0.0145} & 0.6295{\scriptsize $\pm$ 0.0256} & 0.6824{\scriptsize $\pm$ 0.0126} \\            
            GINIPA & DG & 0.7862{\scriptsize $\pm$ 0.0272} & 0.8435{\scriptsize $\pm$ 0.0208} & 0.7676{\scriptsize $\pm$ 0.0122} & 0.7991{\scriptsize $\pm$ 0.0134} \\
                        \midrule
                        
            Tent & TTA & 0.1952{\scriptsize $\pm$ 0.0963} & 0.3426{\scriptsize $\pm$ 0.1302} & 0.1982{\scriptsize $\pm$ 0.0903} & 0.2453{\scriptsize $\pm$ 0.0806} \\
            \midrule
            
            HA\&Tent & DG\&TTA & 0.6698{\scriptsize $\pm$ 0.0296} & 0.7568{\scriptsize $\pm$ 0.0229} & 0.6838{\scriptsize $\pm$ 0.0305} & 0.7035{\scriptsize $\pm$ 0.0191} \\
                
            Ours & DG\&TTA & \textbf{0.8516{\scriptsize $\pm$ 0.0031}} & \textbf{0.8934{\scriptsize $\pm$ 0.0040}} & \textbf{0.8166{\scriptsize $\pm$ 0.0063}} & \textbf{0.8539{\scriptsize $\pm$ 0.0044}} \\
        
            \midrule
            \midrule
            Setting & & \multicolumn{4}{c}{bSSFP$\to$LGE} \\
            \midrule 
            Method & Category & Myo & LV & RV & Average\\
            \midrule
            
            SrcOnly & - & 0.5625{\scriptsize $\pm$ 0.0316} & 0.7046{\scriptsize $\pm$ 0.0392} & 0.6329{\scriptsize $\pm$ 0.0323} & 0.6333{\scriptsize $\pm$ 0.0341} \\
            SAM++ & - & 0.6997{\scriptsize $\pm$ 0.0089} & 0.8773{\scriptsize $\pm$ 0.0014} & 0.8158{\scriptsize $\pm$ 0.0084} & 0.7976{\scriptsize $\pm$ 0.0031} \\
            MinEnt & - & 0.6021{\scriptsize $\pm$ 0.0075} & 0.7486{\scriptsize $\pm$ 0.0051} & 0.6677{\scriptsize $\pm$ 0.0049} & 0.6728{\scriptsize $\pm$ 0.0014} \\
            
            \midrule
            
            HA & DG & 0.7389{\scriptsize $\pm$ 0.0065} & 0.8880{\scriptsize $\pm$ 0.0016} & 0.8509{\scriptsize $\pm$ 0.0028} & 0.8259{\scriptsize $\pm$ 0.0023} \\
            GINIPA & DG & 0.7557{\scriptsize $\pm$ 0.0114} & 0.9021{\scriptsize $\pm$ 0.0022} & 0.8437{\scriptsize $\pm$ 0.0139} & 0.8338{\scriptsize $\pm$ 0.0074} \\
            \midrule
            
            Tent & TTA & 0.5739{\scriptsize $\pm$ 0.0359} & 0.7658{\scriptsize $\pm$ 0.0348} & 0.6260{\scriptsize $\pm$ 0.0355} & 0.6552{\scriptsize $\pm$ 0.0351} \\
            \midrule
            
            HA\&Tent & DG\&TTA & 0.7188{\scriptsize $\pm$ 0.0133} & 0.8765{\scriptsize $\pm$ 0.0050} & 0.7935{\scriptsize $\pm$ 0.0078} & 0.7963{\scriptsize $\pm$ 0.0016} \\
            Ours & DG\&TTA & \textbf{0.7964{\scriptsize $\pm$ 0.0020}} & \textbf{0.9091{\scriptsize $\pm$ 0.0031}} & \textbf{0.8560{\scriptsize $\pm$ 0.0091}} & \textbf{0.8538{\scriptsize $\pm$ 0.0043}} \\
        
            \bottomrule
        \end{tabular}
    }
\end{table*}

\section{Experiments}
\label{sec:Experiments}

We evaluate our method on two public cross-modality MRI datasets, MS-CMRSeg2019~\citep{zhuang2018multivariate,qiu2023myops} and BraTS2018~\citep{menze2014multimodal,lloyd2017high,bakas2018identifying,bakas2017segmentation} with various cross-modal segmentation tasks. Below, we first introduce the dataset information in \sectionref{sec:dataset} and experiment setup in \sectionref{sec:ExperimentSetup}. Then we present our experimental results in \sectionref{sec:results} and ablation study in \sectionref{sec:ablation}.

\subsection{Datasets}
\label{sec:dataset}
\paragraph{The MS-CMRSeg2019 Dataset}~\citep{zhuang2018multivariate,qiu2023myops} features cardiac MRIs from 35 patients across three modalities: balanced Steady-State Free Precession (bSSFP), Late Gadolinium Enhancement (LGE), and T2-weighted MRI. Each patient has multiple slices (typically 8-12 for bSSFP, 10-18 for LGE, and 3-7 for T2). The patients have an average age of $56.2 \pm 7.92$ years and an average weight of $74.4\pm 5.65$ kg \citep{zhuang2022cardiac}. The dataset targets the segmentation of three essential cardiac structures: Myocardium (Myo), Left Ventricle (LV), and Right Ventricle (RV). Given that bSSFP offers clearer visibility for these structures and is preferred for clinical annotations, we designate it as the source domain, and treat LGE and T2 as the target domains~\citep{zhuang2022cardiac}.  


\paragraph{The BraTS2018 Dataset}~\citep{menze2014multimodal,lloyd2017high,bakas2018identifying,bakas2017segmentation} features brain MRI scans of 285 patients, captured across four modalities: T1, T2, post-contrast T1-weighted (T1ce), and T2 Fluid Attenuated Inversion Recovery (FLAIR). The dataset includes annotations across four categories, but we follow previous work~\citep{xie2022unsupervised, zou2020unsupervised, han2021deep}, to focus on the domain generalization of the Whole Tumor (WT) region. Given that the T2 and FLAIR modalities provide clearer visibility of the WT region, we select them as our source domains, and experiment with the domain generalization settings from T2 to T1 and T1ce, as well as from FLAIR to T1 and T1ce.


\subsection{Experiment Setup}
\label{sec:ExperimentSetup}
\paragraph{Baselines} We compare our methods with a diverse set of existing techniques to establish a comprehensive performance benchmark. These include single-domain generalization approaches, test-time adaptation techniques, unsupervised domain adaptation methods, and several other baseline models. Below, we outline the specific methods used for comparison:
(a) \textbf{SrcOnly}, a baseline method that trains models on the source domain without any DG techniques.
(b) \textbf{SAM++~\citep{kirillov2023segment}}, a zero-shot large-scale segmentation model. We use bounding box prompts generated from the prediction of our model. For the myocardium class, we also incorporate a negative point at its geometrical center. 
(c) \textbf{MinEnt~\citep{vu2019advent}}, an Unsupervised Domain Adaptation (UDA) technique that aims to minimize the entropy of target predictions during the training phase. 
(d) \textbf{HA~\citep{vesal2021adapt}}, A DG approach that leverages a variety of pre-defined transformations to augment the training data.
(e) \textbf{GINIPA~\citep{ouyang2022causality}}, a DG method that uses a random weighted network and random spatial interpolation to stylize the image.
(f) \textbf{Tent~\citep{wang2020tent}}, a TTA method that employs entropy minimization during the test stage.
(g) \textbf{HA~\citep{vesal2021adapt}+Tent~\citep{wang2020tent}}, a DG+TTA baseline that integrates HA and Tent for training and test stage.

\paragraph{Evaluation}
Following the literature~\citep{ouyang2022causality,vesal2021adapt,vesal2019automated}, we employ the Volumetric Dice Score as the evaluation metric. we conducted all experiments three times and
report the results with mean and standard derivation.

\paragraph{Implementation Details} 
We employ U-Net architecture~\citep{ronneberger2015u} as the segmentation model. During the training phase, the batch size is $24$, and the learning rate is $0.0001$. All models undergo training for $10000$ iterations. In the training stage, we utilize an online imaging library (imgAug\footnote{\url{https://github.com/aleju/imgaug}}) to implement Heavy Augmentation~\citep{vesal2021adapt}. In the test stage, we adopt horizontal flipping as the Test Time Augmentation function. The parameter $\rho$ remains constant at $0.4$ for all experiments, while $\lambda$ is set to $1$. Additionally, $\alpha$ is configured as $0.9$ for MS-CMRSeg2019 and $0.5$ for BraTS2018. The adaptation learning rate is set at $0.01$, and each image undergoes two adaptation iterations. Moreover, $\mathcal{C}_{lv}$ used for MCE loss is set to be $\{Myo\}$ and $\{WT\}$ for MS-CMRSeg2019 and BraTS2018 respectively. We implemented the code framework via PyTorch on a single NVIDIA A40 (48GB). More implementation details are described in~\appendixref{sec:MoreDetails}.

\subsection{Results}
\label{sec:results}
\paragraph{MS-CMRSeg2019 Results} 
As demonstrated in Table~\ref{tab:comparison_MSCMRSeg2019}, our approach consistently surpasses previous methods across all classes and settings. Its performance is particularly noteworthy in the T2 target domain, where the domain shift from the source domain bSSFP is significant, as visualized in~\figureref{fig:AD_img} (a). Specifically, our method improves the baseline average class Dice score from a mere $4.9\%$ to an impressive 85\%. Furthermore, it outperforms the previous state-of-the-art (SOTA) method, GINIPA~\citep{ouyang2022causality}, by an average of $5.4\%$ in Dice score.

\begin{table*}[h]
    \floatconts
      {tab:comparison_BraTS2018}%
      {\caption{Quantitative comparisons in terms of Dice scores on BraTS2018 dataset.}}%
      {
          \adjustbox{max width=\linewidth}{
              \begin{tabular}{l c c c c}
                \toprule
                
                Setting & T2$\to$T1 & T1$\to$T1ce & Flair$\to$T1 & Flair$\to$T1ce\\
                
                \midrule
                
                SrcOnly & 0.0805{\scriptsize $\pm$ 0.0032} & 0.1528{\scriptsize $\pm$ 0.0134} & 0.0616{\scriptsize $\pm$ 0.0026} & 0.3423{\scriptsize $\pm$ 0.0061} \\
                                
                SAM++ & 0.6475{\scriptsize $\pm$ 0.0024} & 0.6224{\scriptsize $\pm$ 0.0188} & 0.4048{\scriptsize $\pm$ 0.0113} & 0.5492{\scriptsize $\pm$ 0.0079} \\
                                
                MinEnt & 0.0657{\scriptsize $\pm$ 0.0058} & 0.1157{\scriptsize $\pm$ 0.0156} & 0.0755{\scriptsize $\pm$ 0.0022} & 0.3647{\scriptsize $\pm$ 0.0102} \\
                
                \midrule
                
                HA & 0.0778{\scriptsize $\pm$ 0.0017} & 0.1724{\scriptsize $\pm$ 0.0244} & 0.0699{\scriptsize $\pm$ 0.0034} & 0.3630{\scriptsize $\pm$ 0.0027} \\
                                
                GINIPA & 0.2302{\scriptsize $\pm$ 0.0107} & 0.2840{\scriptsize $\pm$ 0.0093} & 0.1865{\scriptsize $\pm$ 0.0118} & 0.3607{\scriptsize $\pm$ 0.0068} \\
                
                \midrule
                
                Tent & 0.0327{\scriptsize $\pm$ 0.0034} & 0.1117{\scriptsize $\pm$ 0.0075} & 0.0463{\scriptsize $\pm$ 0.0045} & 0.2790{\scriptsize $\pm$ 0.0167} \\
                
                \midrule
                
                HA\&Tent & 0.0585{\scriptsize $\pm$ 0.0078} & 0.1371{\scriptsize $\pm$ 0.0183} & 0.1012{\scriptsize $\pm$ 0.0159} & 0.3688{\scriptsize $\pm$ 0.0143} \\
                                
                Ours & \textbf{0.6793{\scriptsize $\pm$ 0.0028}} & \textbf{0.6876{\scriptsize $\pm$ 0.0035}} & \textbf{0.4260{\scriptsize $\pm$ 0.0074}} & \textbf{0.5958{\scriptsize $\pm$ 0.0038}} \\
                
                \bottomrule
            \end{tabular}
          }
      }
\end{table*}
\begin{figure*}[h]
    \floatconts
    {fig:short_comparison}
    {\caption{Comparison of cross-modality segmentation results between baselines and our method under four task settings. Our method achieves superior outcomes in these challenging cases.}\vspace{-5mm}}
    {\includegraphics[width=0.9\linewidth]{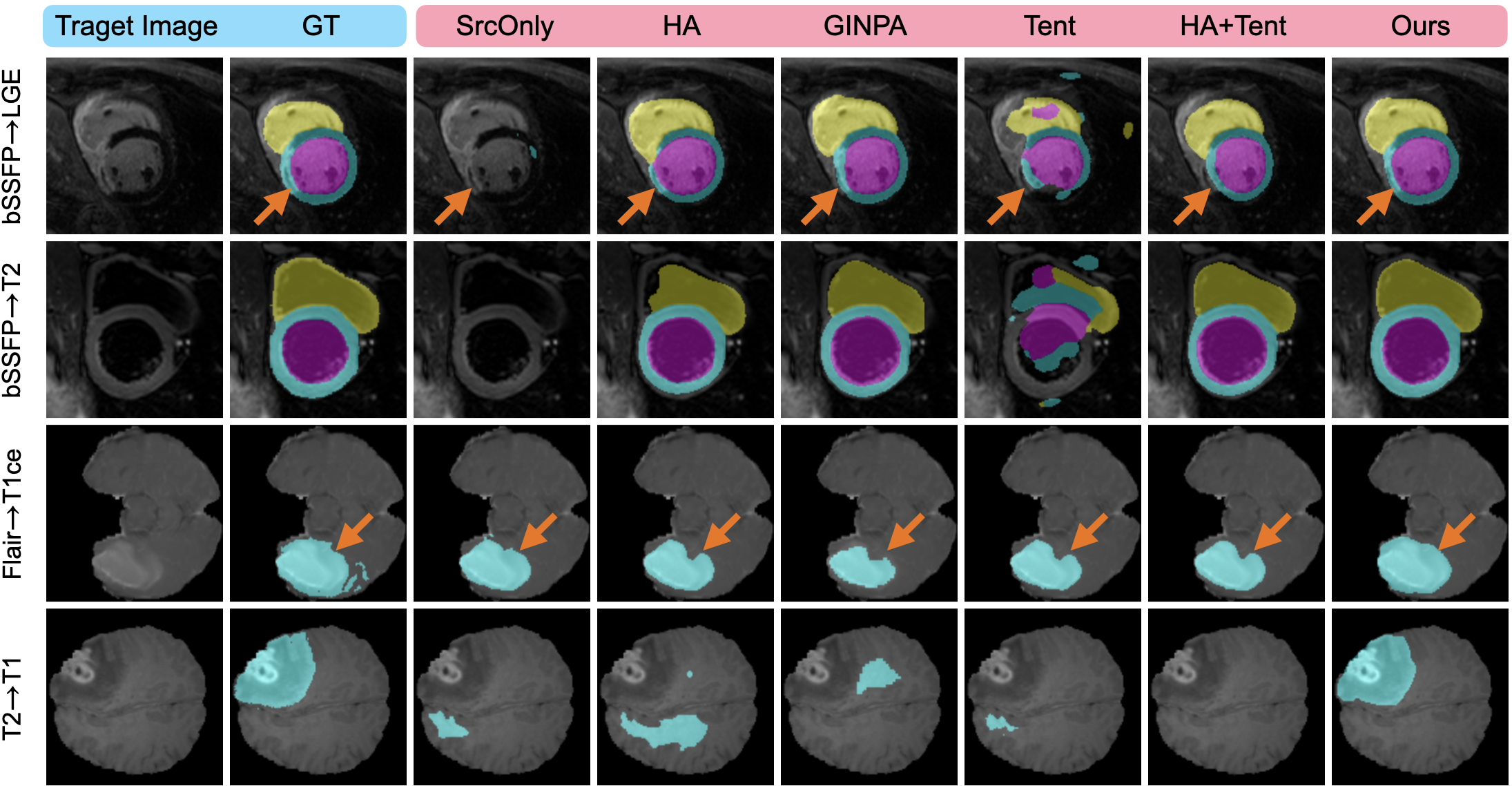}}
\end{figure*}

\paragraph{BraTS2018 Results} Table~\ref{tab:comparison_BraTS2018} showcases the quantitative performance comparison on the BraTS2018 dataset, where our method achieves the best performance for all settings. Interestingly, we observe that most previous methods fall short on this dataset. One possible reason for this could be the pronounced local variations among different modalities, as shown in~\figureref{fig:AD_img} (b). These local variations seem to present difficulties for previous techniques that exclusively depend on image-level augmentations, such as HA\citep{vesal2021adapt} and GINIPA\citep{ouyang2022causality}.


\paragraph{Result Visualization} 
As evidenced by~\figureref{fig:short_comparison}, our method generates segmentation masks that closely resemble the ground truth. Notably, it accurately segments regions impacted by local appearance discrepancies, as indicated by the arrows in the graph. This underscores the effectiveness of our proposed PITTA component in mitigating the effects of local domain shifts. Additional visual results are provided in Appendix~\ref{sec:MoreVisualization},~\figureref{fig:long_comparison}.

\begin{table}[h]
    \floatconts
    {tab:ablation}%
    {\caption{Ablation study on our model components. Performed on MS-CMRSeg2019 Dataset.}}%
    {\adjustbox{max width=\linewidth}{
        \begin{tabular}{c c c c c}
            \toprule
            HA & GMR & PITTA & bSSFP$\to$T2 & bSSFP$\to$LGE\\
            \midrule
            \ding{51} & \ding{55} & \ding{55} & 0.6824{\scriptsize $\pm$ 0.0126} & 0.8259{\scriptsize $\pm$ 0.0023}\\ 
            \ding{51} & \ding{51} & \ding{55} & 0.8430{\scriptsize $\pm$ 0.0066} & 0.8383{\scriptsize $\pm$ 0.0044}\\
            \ding{51} & \ding{55} & \ding{51} & 0.7393{\scriptsize $\pm$ 0.0257} & 0.8400{\scriptsize $\pm$ 0.0012}\\
            \ding{51} & \ding{51} & \ding{51} &\textbf{0.8539{\scriptsize $\pm$ 0.0044}} & \textbf{0.8538{\scriptsize $\pm$ 0.0043}}\\
            \bottomrule
        \end{tabular}
        }
    }
\end{table}

\subsection{Ablation Study}
\label{sec:ablation}
We assess the effectiveness of our proposed architecture by conducting ablation studies on its two main components: GMR and PITTA. Table~\ref{tab:ablation} presents a performance comparison of our method's variants on the MS-CMRSeg2019 dataset under two different settings. In the 2nd row, incorporating the GMR component improves the baseline method (HA) by increasing the average volumetric Dice score by $16\%$ and $1.2\%$ on the two target datasets, respectively. Similarly, in the 3rd row, adding the PITTA component boosts the baseline performance, leading to an average Dice score improvement of $5.7\%$ and $1.4\%$. The final row shows that our complete method, which integrates both components, achieves the best performance, thereby confirming the effectiveness of our design choices.



\begin{figure}[h]
    \floatconts
    {fig:efficiency}
    {\caption{Comparisons on data efficiency of our model with GINIPA under bSSFP $\to$ LGE setting on MS-CMRSeg2019 dataset.}\vspace{-5mm}}
    {\includegraphics[width=\linewidth]{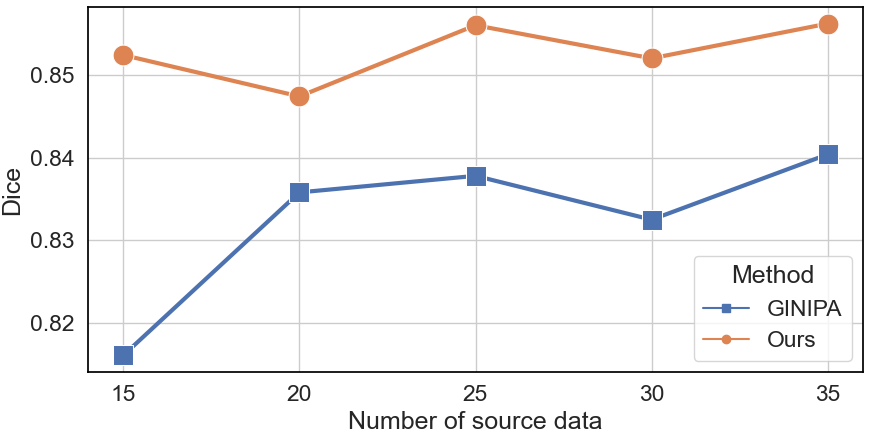}}
\end{figure}

\paragraph{Data Efficiency} 
Given the challenges of obtaining sufficient data and annotations, data-efficiency becomes crucial in medical settings. To evaluate this, we examine the stability of our method under varying training dataset sizes, with the results shown in~\figureref{fig:efficiency}. Specifically, in the MS-CMRSeg2019 dataset, we adjust the number of training patients from the original 35 down to subsets of 15, 20, 25, and 30 and compare our performance to the existing state-of-the-art, GINIPA. Our findings indicate that our model is less sensitive to reductions in dataset size. For instance, when the dataset is trimmed from 35 to 15 patients, the performance of our model declines by only $0.44\%$, compared to a $2.89\%$ drop for GINIPA. This underscores the robustness of our method when data is limited.

For other analysis experiments, please refer to  ~\appendixref{sec:more_quantitative_results} for details.

\section{Conclusion}
\label{Conclusion}
In this paper, we propose a novel adaptive domain generalization framework to address the cross-modal MRI segmentation problem. Our method integrates a learning-free cross-domain representation based on image gradient maps for coping with global domain shift and a class prior-informed test-time adaptation strategy for mitigating local domain shift. We extensively evaluate our method on two multi-modality MRI datasets with six cross-modal segmentation tasks. The results show the superior performance and efficiency of our method in comparison with previous domain generalization and TTA methods. 

\section{Acknowledgements}
This work was supported by Shanghai Science and Technology Program 21010502700, Shanghai Frontiers Science Center of Human-centered Artificial Intelligence, and MoE Key Lab of Intelligent Perception and Human-Machine Collaboration (ShanghaiTech University).
 

\newpage

\bibliography{li23}

\newpage

\appendix

\section{PITTA Pseudo Code}
\label{sec:pseudocode}
\begin{algorithm2e}[htbp]
    \label{alg:pseduocode}
    \caption{Test Procedure of our method}
    \KwIn{input image $\mathbf{x}$, class prior $\mathbf{p}$, $\alpha$, $\lambda$, $\mathcal{C}_{lv}$}
    \KwOut{Segmentation mask $\hat{\mathbf{y}}$}
    \DontPrintSemicolon
    \LinesNumbered
    $\mathbf{q}^{(0)} \gets \frac{\alpha}{C}\mathbf{1}+(1-\alpha)\mathbf{p}^{(0)}$\;
    \For{$n\gets 1, \KwTo N$}{
        $\mathbf{S}^{(n)} \gets \mathcal{M}_{\theta}(G(\mathbf{x}))$\;
        $\mathbf{S}'^{(n)} \gets A_T^{-1}(\mathcal{M}_{\theta}(A_T(G(\mathbf{x}))))$\;
        $L_{cons} \gets |\mathbf{S}^{(n)}-\mathbf{S}'^{(n)}|$\;
        $P^{(n)}(\mathbf{y}|\mathbf{x}) \gets softmax((\mathbf{S}^{(n)}+\mathbf{S}'^{(n)})/2)$\;
        $\hat{\mathbf{y}}^{(n)}_{i,j} \gets \arg\max\left(P^{(n)}(\mathbf{y}_{i,j}|\mathbf{x})/\mathbf{q}^{(n)}\right)$\;
        $\mathbf{M}_{i,j} \gets \bigcup_{c\in \mathcal{C}_{lv}} \mathbf{1}(\mathbf{\hat{\mathbf{y}}}_{i,j}^{(n)} = c)$\;
        $L_{mce} \gets L_{ce}(P^{(n)}(\mathbf{y}|\mathbf{x})\odot \mathbf{M}, \hat{\mathbf{y}}^{(n)})$\;
        $L_{total} \gets L_{mce} + \lambda L_{cons}$\;
        Update $\theta$ using $L_{total}$\;
        $\mathbf{p}^{(n+1)} \gets \alpha \mathbf{p}^{(n)}+(1-\alpha)\#\left\{\hat{\mathbf{y}}^{(n)}\right\}$\;
        $\mathbf{q}^{(n+1)} \gets \frac{\alpha}{C}\mathbf{1}+(1-\alpha)\mathbf{p}^{(n+1)}$\;
    }
    $\hat{\mathbf{y}} \gets \arg\max\left(P^{(N)}(\mathbf{y}|\mathbf{x})/\mathbf{q}^{(N)}\right)$\;
\end{algorithm2e}

\section{More discussion on related work}
\subsection{Our method vs. style-transfer-based techniques}
Most style-transfer-based methods~\citep{tran2019domain,li2017demystifying,atapour2018real} require target-domain images to train the style-transfer function. In contrast, our approach is designed for domain generalization, where no target-domain information is available during training. Essentially, our method could be seen as defining a training-free transfer function via the gradient map, as in Eq.~\eqref{eq:GMR}, to transfer MRI images to a domain-agnostic feature space.

\section{More Implementation Details}
\label{sec:MoreDetails}
\subsection{Data Processing}
For MS-CMRSeg2019 dataset, we resize the LGE and T2 modality to match the shape of bSSFP images and crop them into $128\times 128$. For BraTS2018 dataset, we crop the images via MRI volume bounding box (where intensity $>0$) and resize them into $128\times 128$. Unlike previous protocols~\citep{xie2022unsupervised, zou2020unsupervised, han2021deep} which use random data partitioning, we designate the last 10\% of HGG and LGG data in BraTS2018 as test set to facilitate a fair comparison for the subsequent work.

\subsection{Model Architecture}
We utilize the U-Net~\citep{ronneberger2015u} as our segmentation backbone and adapt the implementation from OpenAI \footnote{\url{https://github.com/openai/guided-diffusion}} in our experiment. Our segmentation model comprises 15 encoder blocks, one middle block, and 15 decoder blocks with different resolutions (from $128\times 128$ to $8\times 8$). The detailed model structure can be found in the official implementation of OpenAI.

\section{Additional Quantitative Results}
\label{sec:more_quantitative_results}
\subsection{Effect of Averaging  Two logits}
\label{sec:average}
In order to demonstrate the effect of taking an average on two logits in Eq.\eqref{eq:posterior}, we empirically compared our method to using outputs from normal images alone. As demonstrated in Table~\ref{tab:avg_compare}, the averaged outputs yield a slight performance improvement. 
\begin{table}[htbp]
\setlength{\tabcolsep}{4pt}
    \floatconts
    {tab:avg_compare}%
    {\caption{A comparison between using only the normal output (w/o prediction averaging) and averaging the outputs from both augmented and normal images (Ours). The experiments were conducted on the MS-CMRSeg2019 dataset, under the bSSFP$\to$LGE setting.}}%
    {\adjustbox{max width=\linewidth}{
        \begin{tabular}{c c c c c}
            \toprule
            & Myo & LV & RV & Average \\
            \midrule
            Normal Output & 0.7961 & \textbf{0.9102} & 0.8594 & 0.8552 \\
            Averaged (Ours) & \textbf{0.7987} & 0.9100 & \textbf{0.8600} & \textbf{0.8562}\\
            \bottomrule
        \end{tabular}
        }
    }
\end{table}

\subsection{Effect of Transformation Type on Test Time Adaptation}
\label{sec:transform_type}
We assessed the performance impact of various transformation functions in test-time augmentation. As shown in Table~\ref{tab:transform_type_compare}, Horizontal Flip is the most efficient for performance improvement in our specific task. 
\begin{table}[htbp]
\setlength{\tabcolsep}{4pt}
    \floatconts
    {tab:transform_type_compare}%
    {\caption{Transformation type comparison conducted on the MS-CMRSeg2019 dataset, under the bSSFP$\to$LGE setting.}}%
    {\adjustbox{max width=\linewidth}{
        \begin{tabular}{c c c c c}
            \toprule
            Transformation types & Myo & LV & RV & Average \\ 
            \midrule
            Rotation 90° & 0.7722 & 0.9040 & 0.8537 & 0.8433 \\
            \midrule
            ColorJitter & 0.7963 & 0.9108 & 0.8473 & 0.8515 \\
            \midrule
            Gaussian Blur & 0.7852 & 0.9089 & 0.8468 & 0.8470 \\
            \midrule
            Horizontal Flip & \textbf{0.7987} & \textbf{0.9100} & \textbf{0.8600} & \textbf{0.8562} \\
            \midrule
            Vertical Flip & 0.7916 & 0.9074 & 0.8565 & 0.8518 \\
            \midrule
            \begin{tabular}[c]{@{}l@{}}Horizontal Flip\\ +ColorJitter\end{tabular} & 0.7980 & 0.9101 & 0.8521 & 0.8534 \\
            \midrule
            \begin{tabular}[c]{@{}l@{}}Horizontal Flip \\ + Vertical Flip\end{tabular} & 0.7956 & 0.9090 & 0.8622 & 0.8556 \\
            \midrule
            \begin{tabular}[c]{@{}l@{}}Horizontal Flip\\ +Rotation $90^{\circ}$\\ +ColorJitter\end{tabular} & 0.7734 & 0.9039 & 0.8442 & 0.8405 \\
            \midrule
            \begin{tabular}[c]{@{}l@{}}Horizontal Flip\\ +Rotation $90^{\circ}$\\ +ColorJitter\\ +Gaussian Blur\end{tabular} & 0.7697 & 0.9060 & 0.8415 & 0.8391 \\ 
            \bottomrule
        \end{tabular}
        }
    }
\end{table}

\subsection{Effect of Pixel Mask}
The pixel mask is selected based on the classes significantly impacted by local variations between domains. In our case, myocardium tissue (Myo) is notably affected due to variations like scar lesions being more visible in LGE scans compared to bSSFP. So we focus on Myo during test time to amplify the effects of these local variations. While domain expertise can be beneficial for pinpointing these key classes, it's not a restriction. If such expertise is lacking, practitioners can opt to include all classes by setting the pixel mask to 'true' for all. As shown in Table~\ref{tab:mask_compare}, this approach results in only a minor performance drop of approximately 0.1\% in the Dice score.

\begin{table}[htbp]
\setlength{\tabcolsep}{2pt}
    \floatconts
    {tab:mask_compare}%
    {\caption{Performance comparison between models using pixel masks (Ours) and without pixel masks.}}%
    {\adjustbox{max width=\linewidth}{
        \begin{tabular}{c c c c c c c}
            \toprule
                Dataset & \multicolumn{4}{c}{BraTS} & \multicolumn{2}{c}{MS-CMRSeg2019} \\
                Setting & T2$\to$T1 & T2$\to$T1ce & Flair$\to$T1 & Flair$\to$T1ce & bSSFP$\to$LGE & bSSFP$\to$2 \\
                Category & WT & WT & WT & WT & Avg & Avg \\
                \midrule
                Ours w/o Mask & 0.6803 & 0.6911 & 0.4091 & 0.5978 & 0.8500 & 0.8433 \\
                Ours & \textbf{0.6813} & \textbf{0.6914} & \textbf{0.4189} & \textbf{0.5986} & \textbf{0.8562} & \textbf{0.8555}\\
            \bottomrule
        \end{tabular}
        }
    }
\end{table}

\subsection{Effect of Regularization Strength}
We assessed the effect of the regularization strength parameter, $\lambda$, on model performance on MS-CMRSeg2019 dataset under bSSFP$\to$T2 setting in Table~\ref{tab:lambda}, revealing that our method is relatively robust to changes in this hyperparameter.

\begin{table}[htbp]
\setlength{\tabcolsep}{4pt}
    \floatconts
    {tab:lambda}%
    {\caption{Effect of $\lambda$ on model performance. Results are averaged over all classes.}}%
    {\adjustbox{max width=\linewidth}{
        \begin{tabular}{c c c c c c c}
            \toprule
            $\lambda$ & 0.1 & 0.3 & 0.5 & 1 & 1.5 & 2 \\
            \midrule
            Avg Dice & 0.7671 & \textbf{0.7738} & 0.7724 & 0.7663 & 0.7617 & 0.7588\\
            \bottomrule
        \end{tabular}
        }
    }
\end{table}

\section{More Result Visualization}
\label{sec:MoreVisualization}
\begin{figure*}[htbp]
    \floatconts
    {fig:long_comparison}
    {\caption{More segmentation visualization between different methods under various settings}}
    {\includegraphics[width=0.7\linewidth]{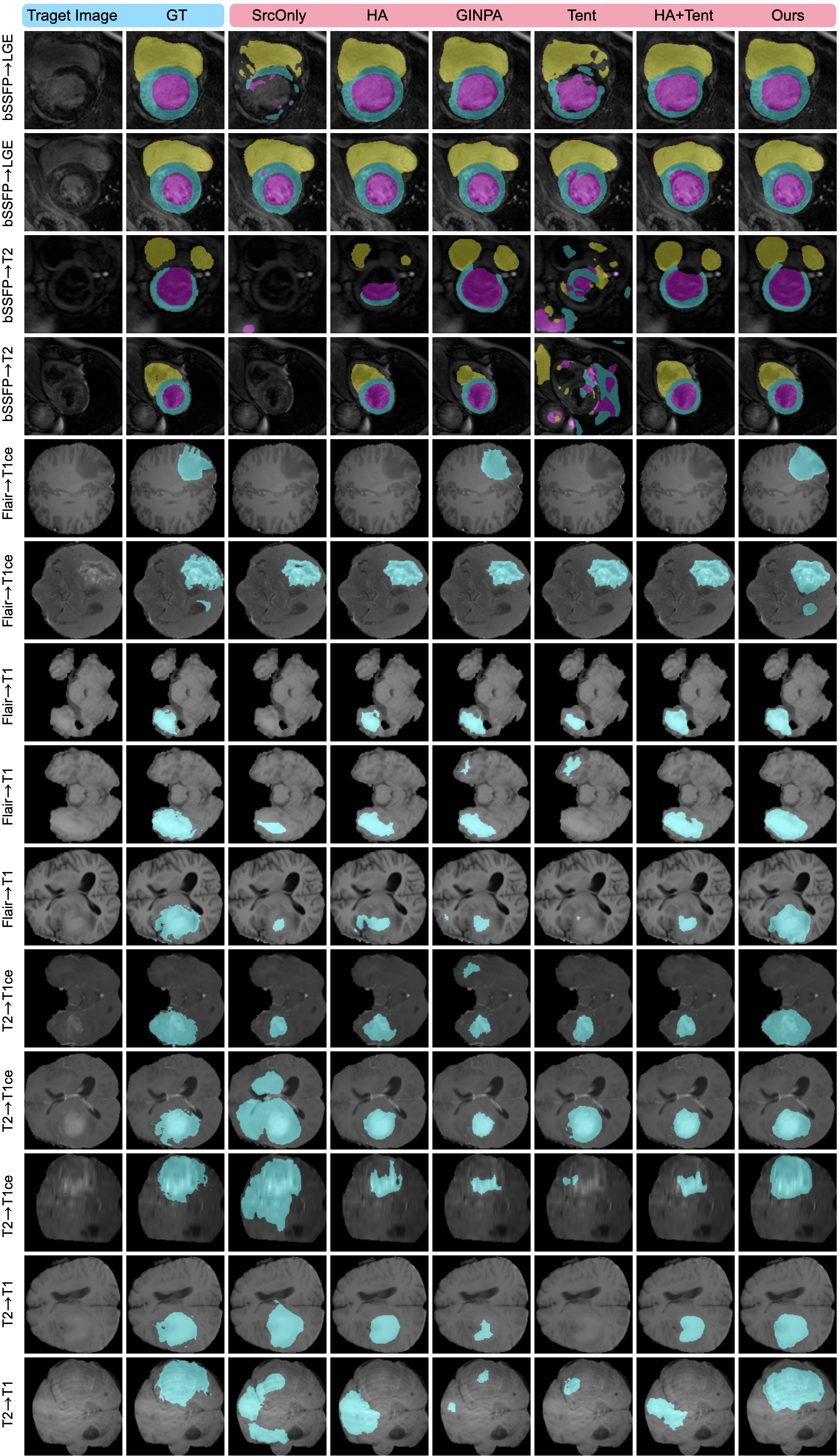}}
\end{figure*}

\end{document}